\documentclass[lettersize,journal]{IEEEtran}
\usepackage{amsmath,amsfonts}
\usepackage{algorithmic}
\usepackage{algorithm}
\usepackage{array}
\usepackage[caption=false,font=normalsize,labelfont=sf,textfont=sf]{subfig}
\usepackage{textcomp}
\usepackage{stfloats}
\usepackage{url}
\usepackage{verbatim}
\usepackage{graphicx}
\usepackage{cite}
\usepackage{booktabs}
\usepackage{multirow}
\usepackage{multicol}
\usepackage{xcolor}
\usepackage{pgfplots}
\pgfplotsset{compat=1.18}

\begin{document}

\title{Fast Cross-Scenario Adaptation of CSI Models via Channel Conditional Parameter Generation}

\author{Xudong Zou, Siyu Wu, Zunlei Feng, Jie Song, Yuanyu Wan, Mingli Song, and Jiacong Hu$^{*}$%
\thanks{All authors are with the School of Software Technology, Zhejiang University, China, and the State Key Laboratory of Blockchain and Data Security, Zhejiang University Hangzhou High-Tech Zone (Binjiang) Institute of Blockchain and Data Security, China. (E-mail: \{xudongzou, wsy041015, zunleifeng, sjie, wanyy, brooksong, jiaconghu\}@zju.edu.cn)}%
\thanks{Corresponding author: Jiacong Hu.}%
\thanks{Code: https://github.com/Z-JiuRi/CCPG}}

\maketitle

\begin{abstract}
The integration of deep learning with massive multiple-input multiple-output (Massive MIMO) systems has shown substantial potential, particularly for physical-layer tasks such as channel state information (CSI) feedback and channel estimation. 
However, severe environmental heterogeneity can substantially degrade CSI model performance in previously unseen scenarios, while conventional adaptation methods often require target-scenario data and considerable computational overhead, thereby hindering fast deployment.
To enable fast adaptation, parameter generation provides a promising paradigm. However, applying it to wireless models remains challenging because channel conditions are high-dimensional and the intrinsic topology of model parameters must be preserved.
To address these challenges, this paper proposes Channel Conditional Parameter Generation (CCPG), an end-to-end conditional parameter generation pipeline for fast deployment of CSI models in dynamic wireless scenarios.
CCPG first identifies scene-sensitive adaptation bottlenecks through component-freezing experiments, allowing parameter generation to focus on lightweight LoRA weights rather than the full CSI model.
It then compresses high-dimensional channel features into compact latent conditions through cascaded SVD and a Perceiver Resampler.
To stabilize the target parameter space, CCPG introduces an energy-based canonicalization mechanism to mitigate permutation and sign ambiguities in LoRA weights.
Finally, the diffusion-based generator incorporates structural information and an asymmetric size-aware loss for topology-aware parameter generation.
Experiments on the DeepMIMO and WAIR-D datasets for CSI feedback and channel estimation demonstrate that CCPG achieves scenario adaptation within approximately 3 seconds through a single forward pass, without training or fine-tuning on new-scenario data, while attaining cross-domain recovery performance comparable to that of costly online adaptation.
These results indicate that CCPG provides an efficient fast-deployment solution for large-scale dynamic wireless scenarios in intelligent 6G communications.

\end{abstract}


\section{Introduction}
\label{sec:intro}

\IEEEPARstart{T}{raditional} signal processing algorithms face increasing difficulty in meeting the stringent requirements of ultra-low latency and massive connectivity in 6G systems when applied to extremely large-scale antenna arrays~\cite{saad2020vision}. As a result, deep learning models have been introduced into physical-layer design and key massive MIMO tasks, such as channel estimation and CSI compression feedback~\cite{ye2018power,11223098,11311690}. Recent 6G studies also explore feedback-free and generative transmission mechanisms to further reduce CSI acquisition overhead~\cite{10750278,11283103}. In practical deployments, the base station (BS) typically maintains a unified and fixed CSI model to support large-scale macroscopic scenarios. However, because of environmental heterogeneity, the CSI feature distribution observed by user equipment (UE) can undergo significant domain shifts as the UE frequently switches among scenarios affected by multipath propagation, building blockage, and variations in channel characteristics. This inherent asymmetry between the fixed CSI model at the BS and the changing channel distribution at the UE can lead to severe generalization degradation and loss of communication performance when the pretrained model is deployed in unseen scenarios. Therefore, enabling fast, flexible, and plug-and-play model adaptation for UEs in different scenarios, while avoiding excessive computational overhead, has become a critical issue for improving the robustness of intelligent physical-layer communication in 6G systems.

To mitigate performance degradation caused by domain shifts, various approaches have been explored in wireless communications, including data-driven methods, model-driven methods, domain adaptation mechanisms, and meta-learning strategies. Data-driven methods attempt to improve generalization by expanding the training set~\cite{wen2018csinet}, but it is difficult for them to exhaustively cover the highly diverse and dynamic potential scenarios. Model-driven methods incorporate domain knowledge to manually design robust features~\cite{he2019modeldriven}, yet they lack dynamic adaptability when facing entirely unseen scenarios. Domain adaptation methods~\cite{feng2023ddanet} promote model transfer by aligning feature distributions, whereas meta-learning~\cite{finn2017model,zhao2025enhanced} aims to learn an initialization that can rapidly converge with only a few target samples. Nevertheless, these conventional cross-scenario adaptation mechanisms still rely on target-scenario data collection and inevitably require multiple online gradient updates, making it difficult to satisfy the practical demand for ready-to-use fast generalization and extremely low deployment cost.

Given the high cost of conventional online training mechanisms, the parameter generation paradigm, which eliminates online gradient updates, provides a promising technical route for fast cross-domain adaptation. In general vision and natural language processing domains, generating lightweight adaptation modules to reduce computational overhead has become an emerging research trend. However, directly transferring existing parameter generation mechanisms to wireless communication systems encounters two fundamental challenges related to condition representation and weight topology. On the one hand, existing generative models heavily rely on low-dimensional and semantically explicit discrete conditions. When applied to high-dimensional and redundant CSI signals, they are prone to the curse of dimensionality and conditional representation collapse. On the other hand, preliminary generation schemes in the communication domain often crudely flatten multi-layer network parameters into one-dimensional vectors, completely discarding the spatial topology of the network architecture as well as the inherent permutation symmetry and sign ambiguity of neural network weights. This may cause mode averaging under complex channel distributions, resulting in generated weights that degenerate into mean parameters lacking physical significance. Therefore, achieving dynamic model adaptation with both robust high-dimensional conditional guidance and structured topological awareness under extremely low deployment overhead remains a key obstacle in the development of intelligent 6G physical-layer systems.

To overcome the dual challenges of condition representation and topology awareness, this paper proposes (CCPG), an end-to-end conditional parameter generation pipeline tailored for asymmetric communication architectures. CCPG addresses the technical bottlenecks of parameter generation in wireless communications and enables fast, plug-and-play model parameter adaptation to unseen communication scenarios.
The overall method consists of four stages. First, during parameter preparation, to avoid the computational burden of full-parameter generation, this paper identify the core bottleneck responsible for cross-scenario generalization through module-level freezing experiments, thereby reducing the dimensionality of the parameters to be generated by several orders of magnitude. Second, during condition extraction, this paper address the curse of dimensionality caused by high-dimensional and redundant CSI signals by designing a feature extraction module based on cascaded SVD and a Perceiver Resampler, which distills high-dimensional channel features containing substantial redundant information into compact and semantically informative conditional latent variables. Third, to resolve the inherent permutation symmetry and sign ambiguity of neural network weights during parameter alignment, this paper introduce an energy-ordering-based canonicalization mechanism to align the weights, establishing a unique and well-defined optimization target for the diffusion backbone and effectively avoiding the mode averaging phenomenon. Finally, during generation, to overcome the topological collapse caused by crudely flattening weights in conventional methods, this paper proposes a weight partitioning strategy, explicitly inject structured positional encodings containing layer-level and matrix-level identifiers, and introduce size-aware loss weighting. These designs enable the diffusion model to accurately perceive and reconstruct the complex inter-layer spatial topology of communication networks during the denoising process. Extensive experiments on realistic and complex MIMO channel estimation and CSI compression feedback tasks demonstrate the superior generalization capability of CCPG. Specifically, CCPG not only substantially reduces the time cost of scenario adaptation exponentially, but also achieves cross-domain recovery performance comparable to expensive target-scenario online training, providing a highly feasible and general solution for large-scale dynamic deployment in intelligent 6G communication systems.

The main contributions of this paper are summarized as follows:
\begin{itemize}

    \item \textbf{A fast-deployment adaptation paradigm for wireless communication models:}
    This paper proposes a conditional parameter generation framework that rapidly adapts the CSI model to newly encountered scenarios. Once trained, the framework directly generates environment-adaptive parameters through a single forward pass, enabling fast, training-free deployment in dynamic wireless scenarios.
    
    \item \textbf{Domain bottleneck localization and lightweight adaptation strategy:}
    Fine-grained component-wise freezing ablations are conducted to localize scene-sensitive modules in communication decoders. Using Transformer-based decoders as a representative case, this paper identifies FFN layers as a major adaptation bottleneck under environmental heterogeneity, motivating lightweight LoRA parameter generation for these critical modules instead of costly full-parameter adaptation.

    \item \textbf{An end-to-end topology-aware generation architecture:} 
    To address the high-dimensional and topological challenges in parameter generation, this paper develops a unified architecture that combines compact condition encoding, energy-based parameter alignment, and structure-aware weight modeling. By preserving the spatial topology of communication weights, the proposed architecture enables accurate generation of adaptive parameters in dynamic wireless scenarios.

    \item \textbf{Superior cross-scenario generalization performance:} 
    Cross-scenario evaluations on DeepMIMO and WAIR-D show that CCPG achieves fast adaptation within only a few seconds for both channel estimation and CSI compression feedback. Without any training or iterative updates on new-scenario data, it attains recovery performance comparable to that of expensive full online training while introducing extremely low inference overhead.
    
\end{itemize}

\section{Related Work}

To enable efficient adaptation of large-scale MIMO systems in dynamic scenarios, the research community has explored various cross-domain generalization mechanisms.
This section first reviews conventional generalization studies in wireless communications, including data augmentation, model-driven designs, domain adaptation, transfer learning, and meta-learning, while highlighting their dependence on target-scenario data or online gradient updates.
It then discusses existing parameter generation attempts in wireless communications, with emphasis on their limitations in deployment assumptions, high-dimensional condition representation, and weight-topology modeling.

\subsection{Conventional Generalization Studies}
\label{subsec:rw_trad_wireless}

In large-scale MIMO systems, existing studies on environmental generalization and adaptation of conventional deep learning models can be broadly categorized into data-driven augmentation, model-driven design, domain adaptation, and meta-learning. Data-driven methods expand the training set to cover more scenarios~\cite{wen2018csinet,transnet}, while robust CSI compression and efficient channel-estimation accelerators improve reliability under noisy or resource-constrained MIMO environments~\cite{11223098,11311690}. However, due to the highly complex electromagnetic propagation characteristics of real wireless scenarios, it is almost impossible to exhaustively enumerate all potential channel states. Model-driven methods exploit physical-layer prior knowledge to constrain feature representations~\cite{he2019modeldriven}, but manually designed features still lack sufficient dynamic adaptability when facing unseen scenarios. Domain adaptation mechanisms align feature distributions between source and target domains~\cite{feng2023ddanet}, whereas meta-learning learns highly generalizable initialization parameters~\cite{finn2017model,zhao2025enhanced}.

Despite these advances, most conventional mitigation methods suffer from two major limitations. First, they heavily rely on continuous offline data collection from target scenarios. Second, they still require multi-step online gradient updates, which hinders truly plug-and-play and efficient environmental adaptation. Therefore, developing fast adaptation schemes that avoid online gradient updates at terminal devices is of both engineering and research significance.

\subsection{Parameter Generation for Generalization}
\label{subsec:rw_wire_param}

In wireless communications, preliminary studies have begun to explore parameter generation techniques for providing environmental adaptability. Current methods mainly use hypernetworks to dynamically generate network parameters for CSI compression feedback~\cite{cen2024hrnet}, channel estimation~\cite{liu2025hypernetwork}, and channel prediction~\cite{hyperchannel2024}. Related 6G feedback-free transmission studies also show the potential of generative modeling for reducing explicit CSI feedback and adapting physical-layer transmission procedures~\cite{10750278,11283103}. By learning a continuous mapping from channel environments to model weights, these methods theoretically endow communication networks with instance-level flexibility. However, they still exhibit fundamental limitations when coping with domain shifts across multiple UEs.

Beyond wireless communications, recent studies increasingly treat neural network parameters as structured objects that can be assembled, diagnosed, repaired, or purified on a parameter manifold~\cite{NEURIPS2024_e6c185ba,NEURIPS2024_614f8eba,NEURIPS2024_5ce377d1,hu2026parameter}. These works suggest the importance of preserving parameter-space structure, but they do not directly address high-dimensional CSI condition representation or asymmetric deployment constraints in communication systems.

First, most existing methods still perform joint parameter updates at both the BS and UE sides, or they rely on UE-side online training during inference to search for optimal conditional inputs~\cite{liu2025hypernetwork}. This substantially conflicts with the practical requirements of asymmetric communication scenarios. Second, existing communication-oriented generation schemes usually adopt simple multilayer perceptrons (MLPs), which are insufficient for extracting informative representations from large-scale CSI data that are often extremely high-dimensional and highly redundant, thereby suffering from the curse of dimensionality. More importantly, these methods directly flatten multi-layer network weights into one-dimensional vectors. Such an operation ignores the spatial topology of network architectures and completely overlooks the inherent permutation symmetry and sign ambiguity of neural network parameters. As a result, the generated weights are prone to mode averaging under complex environmental distributions, leading to severe performance degradation.

Therefore, the field requires a fast-deployment unilateral adaptation scheme that avoids target-scenario training and UE-side online updates. To fill this gap and overcome the dual bottlenecks of high-dimensional condition compression and structured weight generation, this paper constructs an end-to-end CCPG architecture, providing a new technical path for efficient cross-scenario generalization of communication models.

\section{Preliminaries}
\label{sec:preliminaries}

Efficient cross-scenario adaptation of communication models under the stringent computational and latency constraints of terminal devices remains a significant challenge. To clarify the research scope of this problem, this section reviews the neural-network-based communication background and provides a mathematically rigorous definition of the training-free domain generalization problem, thereby establishing the problem framework for the subsequent objective optimization.

\subsection{Background Definition}
\label{subsec:prelim_background}

In Massive MIMO systems, the BS must accurately acquire the mathematical characterization of the electromagnetic propagation path between the transmitter and the receiver, namely the channel state information (CSI), in order to exploit large-scale antenna arrays for high spectral efficiency and spatial multiplexing gains.

In the frequency division duplexing (FDD) mode, the acquisition of CSI gives rise to two core tasks in the wireless physical layer. The first is \textbf{channel estimation}: the BS transmits a pilot matrix $\mathbf{X}$, and the user equipment (UE) receives the signal $\mathbf{Y} = \mathbf{H}\mathbf{X} + \mathbf{N}$, which is attenuated by the channel $\mathbf{H}$ and corrupted by noise $\mathbf{N}$. The UE then reconstructs the high-dimensional channel feature through a parameterized mapping function:
\begin{equation}
    \label{eq:ce_mapping}
    \hat{\mathbf{H}} = \mathcal{F}_{\mathrm{CE}}(\mathbf{Y}; \Theta_{\mathrm{CE}})
\end{equation}
The second is \textbf{CSI compression feedback}: because the uplink bandwidth is limited, the UE compresses the estimated $\hat{\mathbf{H}}$ into a low-dimensional feature matrix $\mathbf{z} = \mathcal{F}_{\mathrm{en}}(\hat{\mathbf{H}}; \Theta_{\mathrm{en}})$ and feeds it back to the BS. The BS then reconstructs the CSI from the received $\mathbf{z}$:
\begin{equation}
    \label{eq:cf_mapping}
    \tilde{\mathbf{H}} = \mathcal{F}_{\mathrm{de}}(\mathbf{z}; \Theta_{\mathrm{de}})
\end{equation}

With the development of deep learning techniques, the above two tasks differ in their physical deployment: CSI compression feedback follows a distributed end-to-end transmission paradigm, whereas channel estimation is usually formulated as a local signal mapping problem at the receiver. Nevertheless, mainstream solutions can be uniformly abstracted at the network-topology level as Transformer-based encoder-decoder architectures. Specifically, both $\mathcal{F}_{\mathrm{CE}}$, which extracts features from the noisy signal $\mathbf{Y}$, and $\mathcal{F}_{\mathrm{en}}$, which compresses $\hat{\mathbf{H}}$, can be regarded as implicit encoding processes for feature extraction and dimensionality reduction. Meanwhile, $\mathcal{F}_{\mathrm{de}}$ at the BS acts as a decoder responsible for high-fidelity reconstruction of the target communication data.

\begin{figure}[htbp]
    \centering

    \includegraphics[width=1.0\columnwidth]{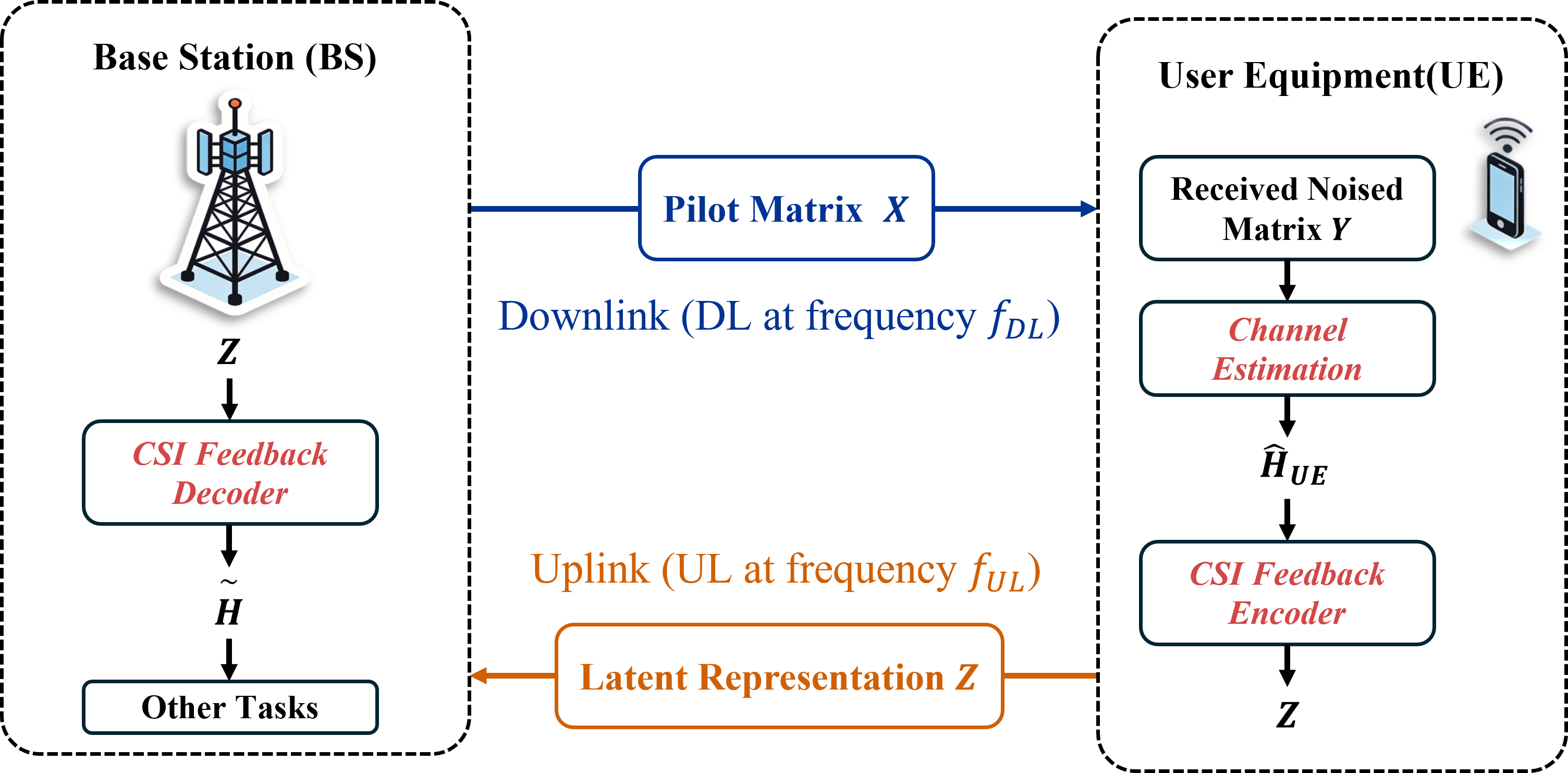}

    {\scriptsize (a)}


    \includegraphics[width=1.0\columnwidth]{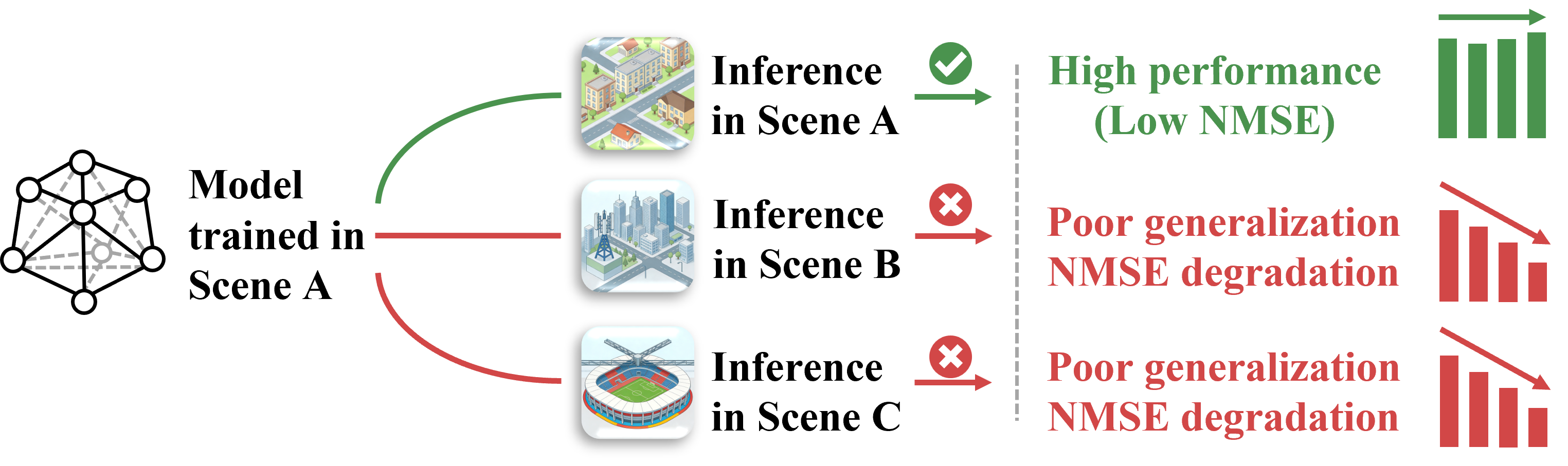}

    {\scriptsize (b)}

    \caption{System model of CSI feedback and the motivation for cross-scenario adaptation. (a) Downlink and uplink transmission procedures between the BS and UE, illustrating the CSI acquisition and feedback loop. (b) Performance degradation of a deep learning-based model when deployed in mismatched environmental scenarios, highlighting the generalization bottleneck.}
    \label{fig:fig1}
\end{figure}

However, in practical heterogeneous networks, when the UE moves from a source scenario to an unseen target scenario, as shown in Fig.~\ref{fig:fig1}, drastic environmental changes may induce severe domain shift in the channel data distribution, i.e., $P_{\mathrm{source}}(\mathbf{H}) \neq P_{\mathrm{target}}(\mathbf{H})$. This inherent architectural consistency of deep learning models, together with the dynamic variation in data distributions, provides the theoretical foundation for this paper to perform lightweight parameter generation on the decoder side for addressing the cross-domain adaptation problem.

\subsection{Problem Formulation}
\label{subsec:prelim_problem_formulation}

Based on the unified encoder-decoder architecture introduced in Section~\ref{subsec:prelim_background}, this paper formulates cross-scenario adaptation of communication models as a rigorous training-free domain generalization problem. Let $\mathcal{S}=\{\mathcal{S}_1,\dots,\mathcal{S}_N\}$ denote a set of $N$ physical scenarios in the network space. For any scenario $\mathcal{S}_i$, its channel state $\mathbf{H}$ and the corresponding network input signal $\mathbf{s}$ follow a specific joint probability distribution $P_{\mathcal{S}_i}(\mathbf{s},\mathbf{H})$. Because of environmental heterogeneity, significant distribution shifts exist across different scenarios, namely $P_{\mathcal{S}_i} \neq P_{\mathcal{S}_j}$, which causes a static model fitted on a single scenario to suffer severe communication performance degradation when deployed in a new scenario. This paper defines the first $N-1$ visible scenarios as the source-domain set $\mathcal{D}_{\mathrm{src}}=\{\mathcal{S}_i\}_{i=1}^{N-1}$, and the completely unseen $N$-th scenario as the target domain $\mathcal{S}_N$.

Under the ideal online training paradigm, to eliminate performance degradation, the model needs to collect environmental data and perform backpropagation after being deployed to any scenario $\mathcal{S}_i$, so as to solve for the optimal decoder adaptation parameters $\mathbf{W}_i^*$ for that specific domain through empirical risk minimization:
\begin{equation}
    \label{eq:oracle_weight}
    \begin{split}
        \mathbf{W}_i^* = \arg\min_{\mathbf{W}}\mathbb{E}_{(\mathbf{s},\mathbf{H})\sim P_{\mathcal{S}_i}}\Bigg[ &
        \mathcal{L}_{\mathrm{task}}\left(\mathbf{H},\right. \\
        & \left.\mathcal{F}_{\mathrm{de}}\left(\mathcal{F}_{\mathrm{en}}(\mathbf{s};\Theta_{\mathrm{en}});\mathbf{W},\Phi\right)\right)\Bigg]
    \end{split}
\end{equation}
where $\mathcal{F}_{\mathrm{en}}$ and $\mathcal{F}_{\mathrm{de}}$ denote the encoder and decoder, respectively; $\Phi$ denotes the remaining frozen decoder parameters; and $\mathbf{W}$ denotes the core bottleneck weights to be optimized. However, on computation-constrained terminal devices, it is highly impractical to frequently perform online gradient updates for the rapidly changing target scenario $\mathcal{S}_N$.

To this end, this paper aims to learn a lightweight conditional parameter generator $\mathcal{G}_\omega$. Under the strict constraint of no online training, the generator relies only on the hidden feature $\mathbf{z} = \mathcal{F}_{\mathrm{en}}(\mathbf{s};\Theta_{\mathrm{en}})$ produced by the fixed encoder, extracts a  low-dimensional scene-semantic condition $\mathbf{c}\in\mathbb{R}^{M\times d}$ as the guidance signal, and directly maps it to the adaptation residual term of the target domain through a single forward pass:
\begin{equation}
    \label{eq:parameter_generator}
    \Delta\mathbf{W}^{(N)} = \mathcal{G}_\omega(\mathbf{c}^{(N)})
\end{equation}
Here, $\Delta\mathbf{W}^{(N)}$ denotes the target-domain residual bottleneck parameters, which are instantiated as LoRA matrices in Section~\ref{subsec:parameter_bottleneck_analyzer}, and $\mathbf{W}_0$ denotes the corresponding frozen pretrained bottleneck weights.

Accordingly, the target-domain generalization objective of this paper can be rigorously defined as finding the optimal generator parameter $\omega$ such that the expected generalization risk $\mathcal{R}_{\mathrm{tgt}}(\omega)$ on the completely unseen target domain $\mathcal{S}_N$ is minimized:
\begin{equation}
\label{eq:target_generalization_obj}
\begin{aligned}
\min_{\omega}\mathcal{R}_{\mathrm{tgt}}(\omega)
=&\;
\mathbb{E}_{(\mathbf{s},\mathbf{H})\sim P_{\mathcal{S}_N}}
\Big[
\mathcal{L}_{\mathrm{task}}
\Big(
\mathbf{H},  \\
&\quad
\mathcal{F}_{\mathrm{de}}
\Big(
\mathbf{z}^{(N)};
\mathbf{W}_0+\mathcal{G}_{\omega}(\mathbf{c}^{(N)}),
\Phi
\Big)
\Big)
\Big].
\end{aligned}
\end{equation}

Clearly, since the distribution of the target domain $\mathcal{S}_N$ is strictly unseen during training, directly optimizing the above integral objective is mathematically intractable. Therefore, this paper performs an \textbf{objective transformation}: instead of directly computing the reconstruction loss in an unknown environment, a conditional manifold diffusion objective is constructed in the high-dimensional parameter space. Specifically, during the offline stage, the CCPG architecture fully exploits the visibility of the source domains $\mathcal{D}_{\mathrm{src}}$ and independently trains the optimal parameter set $\{\mathbf{W}_i^*\}_{i=1}^{N-1}$ for a large number of scenarios according to Eq.~\eqref{eq:oracle_weight}. Subsequently, a meta-dataset composed of paired samples $\langle \text{encoder-guided condition } \mathbf{c}_i, \text{optimal weight } \mathbf{W}_i^* \rangle$ is constructed. Through the structured denoising process of the diffusion model, CCPG learns and generalizes this universal mapping manifold, thereby indirectly and efficiently approximating the theoretical generalization boundary defined in Eq.~\eqref{eq:target_generalization_obj}.

\section{Methodology}
\label{sec:method}

\subsection{Overview}
\label{subsec:overview}

\begin{figure*}[htbp]
    \centering
    \includegraphics[width=1\linewidth]{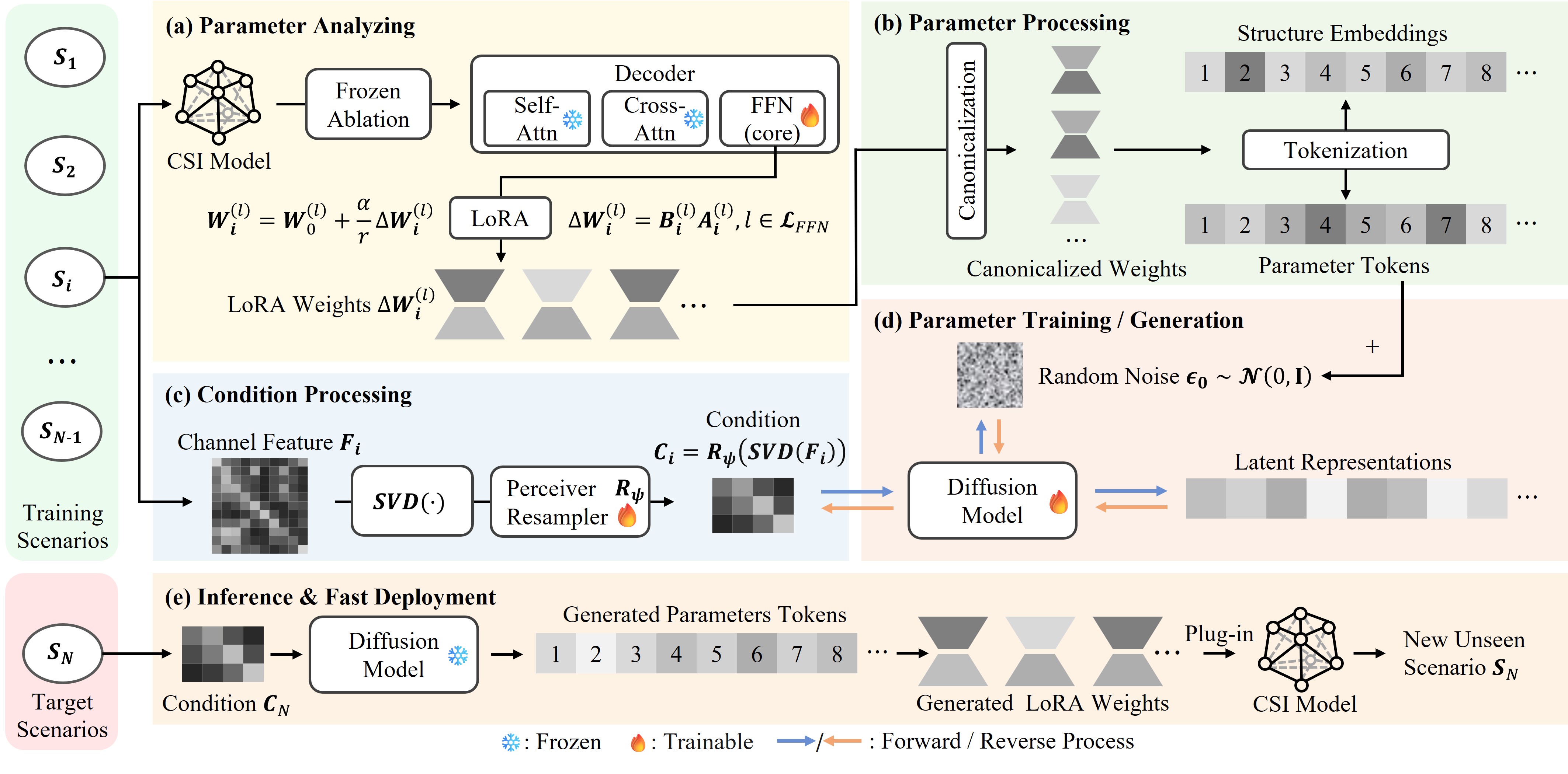} 
    \caption{Overall workflow of the proposed CCPG framework. During offline training, CCPG constructs paired condition-parameter samples from multiple source scenarios, learns a conditional diffusion model over canonicalized LoRA parameter tokens, and then directly generates plug-in adaptation weights for a new target scenario at deployment time. (a) Parameter analyzing localizes the scenario-sensitive decoder component through frozen ablation and selects the FFN LoRA weights as the generation target. (b) Parameter processing canonicalizes the LoRA weights to remove permutation and sign ambiguities and tokenizes them with structure embeddings. (c) Condition processing extracts compact scenario conditions from channel features using SVD and the Perceiver Resampler. (d) Parameter training/generation learns the conditional diffusion process from random noise and scene conditions to parameter tokens. (e) Inference and fast deployment generate target-scenario LoRA weights from the target condition and plug them back into the original CSI model.}
    \label{fig:architecture}
\end{figure*}

To address the training-free domain generalization problem defined in Section~\ref{subsec:prelim_problem_formulation}, this paper proposes CCPG, an end-to-end conditional parameter generation pipeline tailored for asymmetric communication architectures. The overall framework is illustrated in Fig.~\ref{fig:architecture}. The following subsections describe the four key components of the proposed pipeline: the parameter bottleneck analyzer in Section~\ref{subsec:parameter_bottleneck_analyzer}, which localizes domain-specific target parameters; the robust condition encoder in Section~\ref{subsec:robust_condition_encoder}, which extracts compact scene-level semantics; the energy-driven alignment operator in Section~\ref{subsec:canonicalization}, which eliminates ambiguities in the weight space; and the structure-aware diffusion backbone in Section~\ref{subsec:method_generation}, which reconstructs the parameter manifold with high fidelity.

\subsection{Parameter Bottleneck Analysis}
\label{subsec:parameter_bottleneck_analyzer}

Considering the stringent constraints on computation, storage, and latency at terminal devices, generating the full set of decoder parameters is infeasible. For both channel estimation and CSI compression feedback, although the specific network structures and task objectives may differ, the corresponding models can generally be regarded as being composed of multiple functional parameter modules. The performance degradation caused by domain shifts is often concentrated in a small number of critical parameter modules. Therefore, to enable efficient cross-scenario adaptation, the primary objective of this module is to identify the common core parameter bottleneck shared by the two tasks when coping with domain shifts.

\begin{figure}[htbp]
    \centering

    \pgfplotsset{compat=1.18}

    \definecolor{color1}{RGB}{200,213,236}
    \definecolor{color2}{RGB}{254,243,229}
    \definecolor{color3}{RGB}{225,239,216}
    \definecolor{color4}{RGB}{249,218,222}
    
    \begin{tikzpicture}
        \begin{axis}[
            ybar=1pt, 
            bar width=7.5pt,
            width=1.0\columnwidth,
            height=0.65\columnwidth,
            ylabel={NMSE (dB)},
            xlabel={Scenario ID},
            symbolic x coords={09493, 09513, 09614, 09846, 09957},
            xtick=data,
            ymin=-22, ymax=-7,
            enlarge x limits=0.15,
            label style={font=\scriptsize},
            tick label style={font=\scriptsize},
            legend style={
                at={(0.5,-0.25)}, 
                anchor=north,
                legend columns=4, 
                font=\scriptsize,
                draw=gray!80,
                fill=white,
                /tikz/every even column/.append style={column sep=0.2cm}
            },
            grid=major,
            grid style={dashed, gray!30}, 
            axis line style={black},
            nodes near coords,
            every node near coord/.append style={
                font=\scriptsize,
                rotate=90,
                anchor=west,
                check for zero/.code={ 
                    \pgfkeys{/pgf/fpu}
                    \pgfmathparse{\pgfplotspointmeta < -100}
                    \pgfkeys{/pgf/fpu=false}
                },
                /pgf/number format/fixed,
                /pgf/number format/precision=1
            },
        ]

        \addplot[fill=color1, draw=gray!80] 
        coordinates {(09493, -20.2) (09513, -19.8) (09614, -19.5) (09846, -18.9) (09957, -19.8)};
        \addlegendentry{None}

        \addplot[fill=color2, draw=gray!80] 
        coordinates {(09493, -14.5) (09513, -15.8) (09614, -14.8) (09846, -15.2) (09957, -15.2)};
        \addlegendentry{Self-Attn}

        \addplot[fill=color3, draw=gray!80] 
        coordinates {(09493, -15.5) (09513, -15.1) (09614, -14.7) (09846, -15.0) (09957, -14.0)};
        \addlegendentry{Cross-Attn}

        \addplot[fill=color4, draw=gray!80] 
        coordinates {(09493, -13.6) (09513, -12.2) (09614, -10.9) (09846, -10.7) (09957, -11.5)};
        \addlegendentry{FFN}

        \end{axis}
    \end{tikzpicture}
    \caption{NMSE performance comparison under different component-freezing settings for TransNet across multiple scenarios.}
    \label{fig:frozen_components}
\end{figure}
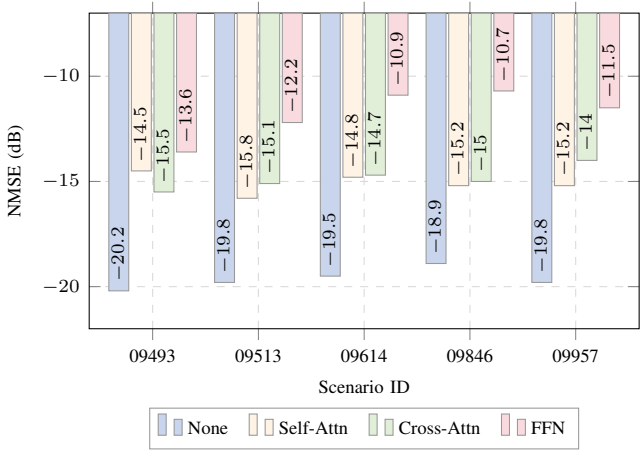

To verify the effectiveness of the parameter bottleneck analyzer in a concrete task, this paper uses TransNet~\cite{transnet}, a representative baseline model for CSI compression feedback, and the public WAIR-D dataset as an example, and conducts module-level freezing cross-scenario ablation experiments on the decoder. TransNet is a classic Transformer-based CSI compression feedback model with a clear modular structure, making it suitable for analyzing the impact of different parameter components on cross-scenario adaptation. As shown in Fig.~\ref{fig:frozen_components}, freezing the feed-forward network (FFN) layer leads to the most severe degradation in reconstruction performance in new scenarios. This empirical result supports the architectural hypothesis that, in Transformer-based models, the attention mechanism mainly focuses on global contextual feature aggregation, whereas the FFN layer acts as a knowledge repository that stores scene-specific nonlinear mappings.

After identifying the FFN as a common bottleneck, this paper further introduces the LoRA mechanism to uniformly perform lightweight adaptation on the FFN linear layers in the decoder, so as to reduce the parameter dimensionality to an acceptable range. Let the original pretrained weight of an FFN layer be $\mathbf{W}_0 \in \mathbb{R}^{d_{\mathrm{out}} \times d_{\mathrm{in}}}$. Its adaptation weight $\mathbf{W}$ under a specific target scenario is parameterized as:
\begin{equation}
    \label{eq:lora_parameterization}
    \mathbf{W} = \mathbf{W}_0 + \mathbf{B}\mathbf{A}
\end{equation}
where the low-rank matrices $\mathbf{A} \in \mathbb{R}^{r \times d_{\mathrm{in}}}$ and $\mathbf{B} \in \mathbb{R}^{d_{\mathrm{out}} \times r}$ satisfy $r \ll \min(d_{\mathrm{in}}, d_{\mathrm{out}})$.

Therefore, for both channel estimation and CSI compression feedback, the optimization space of CCPG for cross-scenario adaptation is transformed from large and redundant full decoder weights into the generation of an extremely lightweight residual parameter set $\mathcal{W}_{\mathrm{LoRA}} = \{\mathbf{A}, \mathbf{B}\}$. While preserving high-quality performance recovery, this analysis and localization strategy mitigates the curse of dimensionality at its source, thereby paving the way for ultra-low-latency inference at terminal devices.

\subsection{Robust Condition Encoder}
\label{subsec:robust_condition_encoder}

In the conditional generation pipeline, the robust condition encoder $\mathcal{E}_{\mathrm{cond}}$ maps high-dimensional and redundant communication hidden features into compact generative priors. Since a large number of user equipment devices within the same macroscopic scenario share similar scattering clusters and macroscopic electromagnetic propagation characteristics, the hidden features $\mathbf{z} = \mathcal{F}_{\mathrm{en}}(\mathbf{s}; \Theta_{\mathrm{en}})$ extracted by the preceding encoder exhibit significant low-rank redundancy. Directly using these features as diffusion conditions may easily lead to the curse of dimensionality in the generation space and fail to provide effective conditional guidance. To avoid introducing additional heavy pretrained components such as VAEs, this paper designs a cascaded condition compression operator following a ``physical truncation $\to$ semantic resampling'' pipeline, thereby achieving a dimensional transition from complex physical signals to compact scene semantics $\mathbf{c}$.

\begin{figure}[htbp]
    \centering

    \includegraphics[width=1.0\columnwidth]{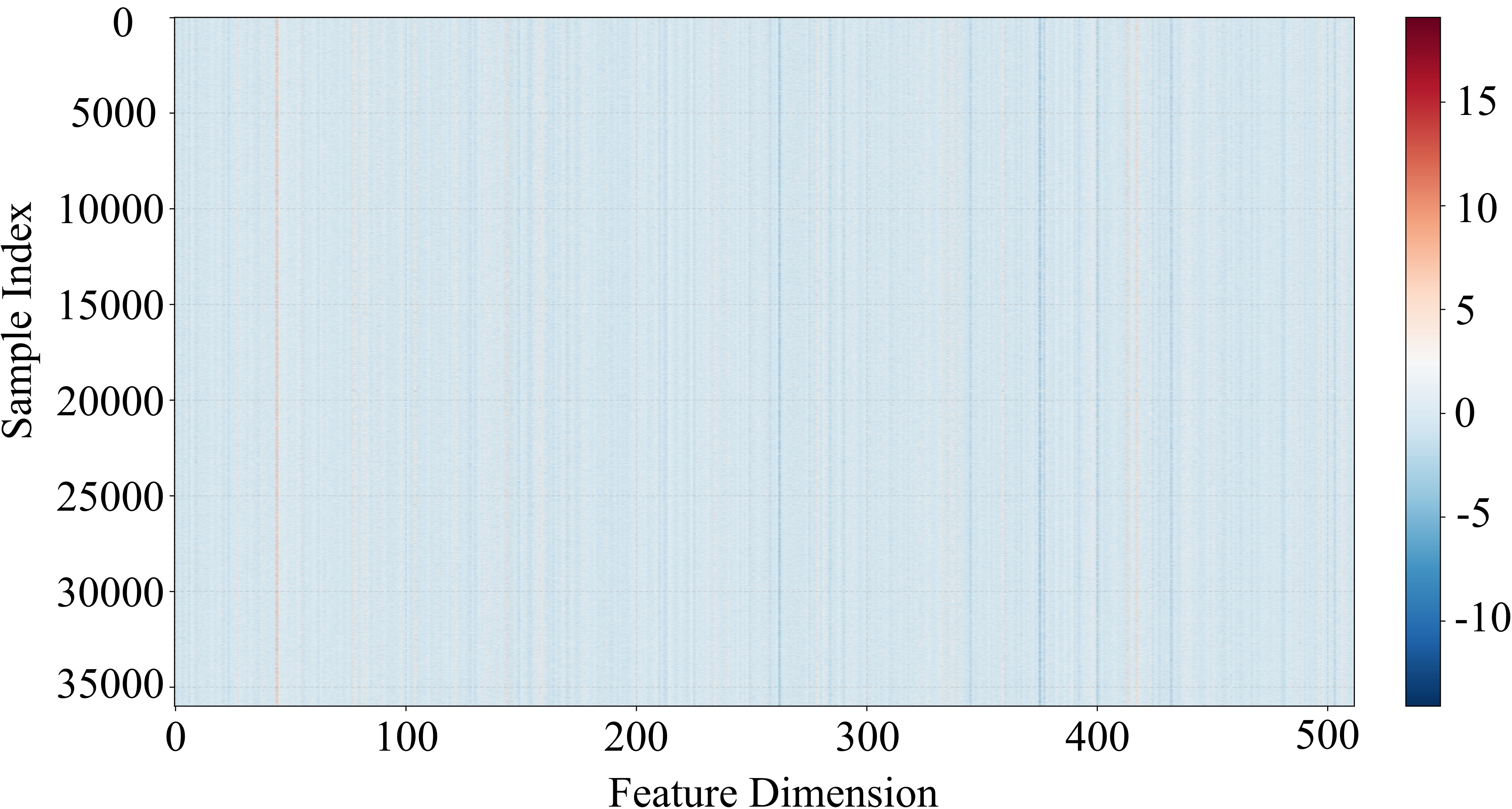}

    \vspace{0.03cm}
    {\scriptsize (a)}

    \vspace{0.18cm}

    \includegraphics[width=1.0\columnwidth]{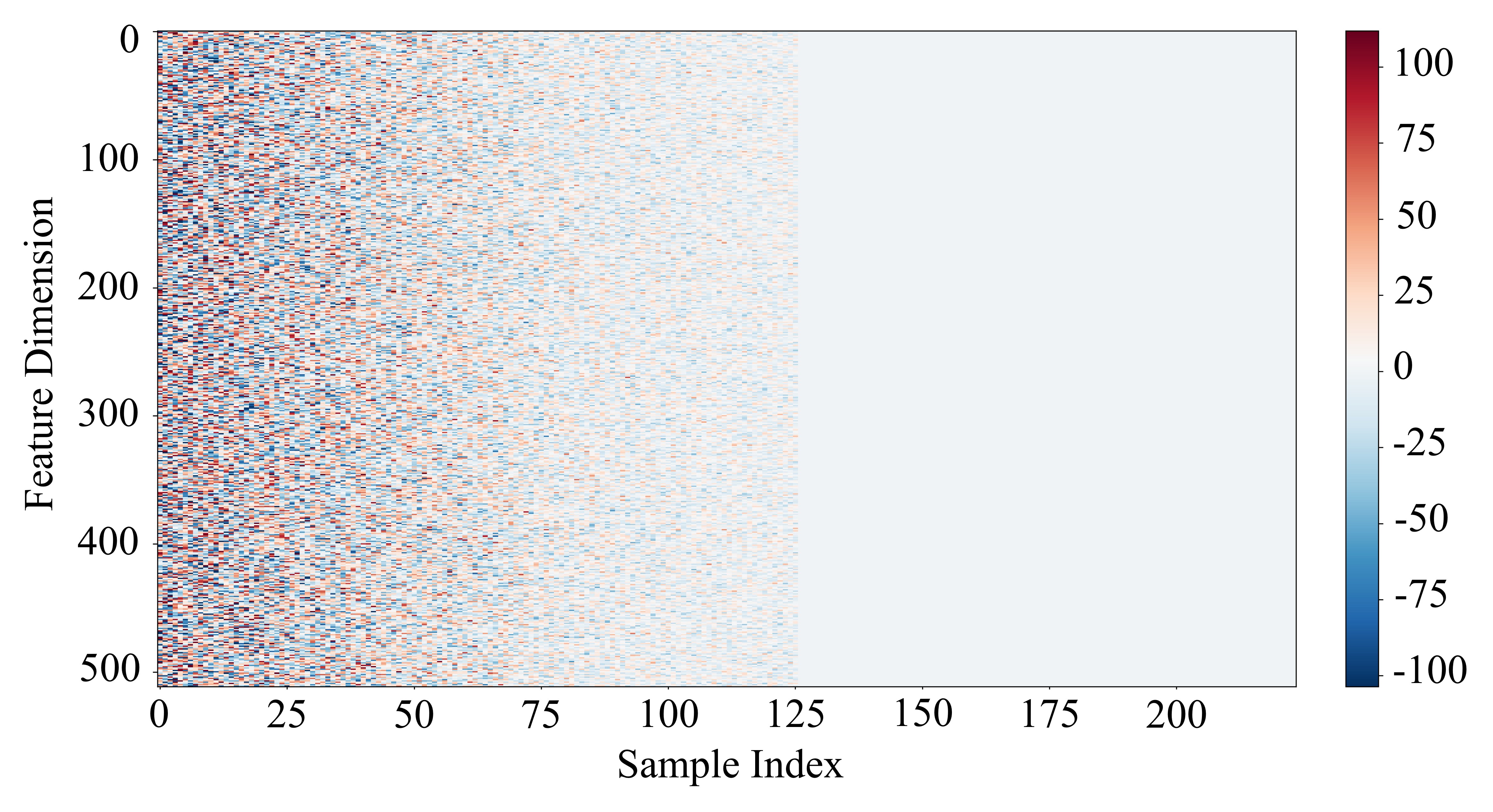}

    \vspace{0.03cm}
    {\scriptsize (b)}

    \caption{Comparison of the channel feature before and after SVD compression. (a) Original high-dimensional feature. (b) SVD-compressed latent feature.}
    \label{fig:svd_comparison}
\end{figure}

First, for high-dimensional communication hidden features, this paper introduces a non-learnable singular value decomposition (SVD) operator for deterministic physical truncation. For the received channel representation matrix $\mathbf{z}\in\mathbb{R}^{n\times d}$, let $\mathbf{z}=\mathbf{U}\boldsymbol{\Sigma}\mathbf{V}^{T}$ denote its SVD. Its first $L$ principal components are extracted to construct the dimension-reduced physical matrix~\cite{eckart1936approximation} $\mathbf{C}_{\mathrm{svd}} = \boldsymbol{\Sigma}_L \mathbf{V}_L^T \in \mathbb{R}^{L \times d}$, where $\boldsymbol{\Sigma}_L$ contains the largest $L$ singular values and $\mathbf{V}_L$ contains the corresponding right singular vectors, as shown in Fig.~\ref{fig:svd_comparison}. From a physical perspective, this matrix preserves the dominant features of the current scenario while avoiding latent-space feature entanglement commonly observed in conventional autoencoders, with no additional training cost.

Second, to meet the fixed-length condition requirement of the diffusion backbone, this paper introduces a Perceiver Resampler jointly trained with the diffusion backbone~\cite{jaegle2021perceiver} to perform cross-dimensional semantic resampling on $\mathbf{C}_{\mathrm{svd}}$. A set of $M$ learnable latent queries $\mathbf{Q}_{\mathrm{latent}} \in \mathbb{R}^{M \times d}$ is defined, where $d$ denotes the latent feature dimension. Through the cross-attention mechanism, the model uses $\mathbf{Q}_{\mathrm{latent}}$ as query anchors and $\mathbf{C}_{\mathrm{svd}}$ as keys and values, filtering out invalid padding and redundant information:
\begin{equation}
    \label{eq:condition_resampling}
    \mathbf{c} = \operatorname{Softmax}\left(\frac{\mathbf{Q}_{\mathrm{latent}} (\mathbf{C}_{\mathrm{svd}}\mathbf{W}_{K})^T}{\sqrt{d}}\right) (\mathbf{C}_{\mathrm{svd}}\mathbf{W}_{V})
\end{equation}
where $\mathbf{W}_{K} \in \mathbb{R}^{d \times d}$ and $\mathbf{W}_{V} \in \mathbb{R}^{d \times d}$ denote the learnable linear projection matrices for keys and values, respectively, while $\mathbf{Q}_{\mathrm{latent}}$ is directly used as the query set. After this mapping, the output $\mathbf{c} \in \mathbb{R}^{M \times d}$ converts variable-length physical features into a compact feature sequence of fixed size $M$.

Thus, the originally high-dimensional, variable-length, and redundant communication feature $\mathbf{z}$ is projected into a unique and compact low-dimensional conditional manifold $\mathbf{c} = \mathcal{E}_{\mathrm{cond}}(\mathbf{z})$. This robust conditional representation decouples the input complexity and reduces the fitting difficulty of the generative model. It also provides smooth semantic guidance for the subsequent diffusion backbone to perform topology-aware generation, thereby mitigating the representation bottleneck on the condition-injection side.

\subsection{Energy-Driven Alignment Operator}
\label{subsec:canonicalization}

When constructing the offline meta-dataset with different random seeds, directly using the extracted domain-specific LoRA weight pair $(\mathbf{A}, \mathbf{B})$ as the fitting target of the diffusion backbone may result in severe representation collapse. Specifically, since the forward mapping is given by $\Delta \mathbf{W} = \mathbf{B}\mathbf{A} = \sum_{i=1}^{r} \mathbf{b}_i \mathbf{a}_i^T$, this lightweight parameter space inherently suffers from permutation symmetry and sign ambiguity~\cite{entezari2022role,ainsworth2023git}. Mathematically, this means that for any permutation matrix $\mathbf{P}\in\{0,1\}^{r\times r}$ with exactly one nonzero entry in each row and column, and any diagonal sign matrix $\mathbf{S}=\operatorname{diag}(s_1,\ldots,s_r)$ with $s_i\in\{-1,1\}$, the identity $\Delta \mathbf{W} \equiv (\mathbf{B}\mathbf{P}\mathbf{S})(\mathbf{S}\mathbf{P}^T\mathbf{A})$ always holds. Such an ill-posed one-to-many mapping causes the diffusion model to average equivalent but spatially divergent weights in the coordinate space during training, resulting in mode averaging and eventually generating smoothed mean parameters without physical significance. To address this issue, this paper proposes a joint energy-constrained operator $\mathcal{O}$, detailed in Algorithm~\ref{alg:canonicalization}, to establish a deterministic canonical transformation for the permutation and sign degrees of freedom. The remaining continuous LoRA scaling freedom is suppressed in practice by fixed LoRA scaling and the global standardization applied after canonicalization.

\textbf{1) Permutation Alignment via Energy Sorting:} The summation order of LoRA rank components is inherently permutable. To establish a deterministic ordering, this paper defines a permutation operation $\pi$ based on the energy of each rank component, where $e_i = \|\mathbf{a}_i\|_2 \cdot \|\mathbf{b}_i\|_2$, such that $e_{\pi(1)} \geq e_{\pi(2)} \dots \geq e_{\pi(r)}$ with the original rank index used as a tie breaker. The corresponding orthogonal permutation matrix $\mathbf{P}^*$ is then constructed accordingly. From a physical perspective, this forces the generative network to prioritize the principal subspaces that contribute the most to channel feature reconstruction during denoising. After applying this transformation, the intermediate matrices with eliminated permutation ambiguity are obtained as $\mathbf{A}' = {\mathbf{P}^*}^T \mathbf{A}$ and $\mathbf{B}' = \mathbf{B} \mathbf{P}^*$.

\textbf{2) Sign Alignment via Anchor Flipping:} After fixing the rank ordering, the row and column vectors corresponding to the same rank can still be simultaneously sign-flipped while preserving the final mapping. To anchor a unique directional manifold, this paper extracts the sign of the pivot element, namely the element with the largest absolute value, in each row of $\mathbf{A}'$ as the reference direction of that vector. Based on this rule, a diagonal sign matrix $\mathbf{S}^*$ is defined. Its diagonal element is given by $s_i = \operatorname{sgn}(\mathbf{A}'_{i, j_i^*})$, where $j_i^* = \arg\max_{j} |\mathbf{A}'_{i,j}|$ and $\operatorname{sgn}(0)$ is set to $1$.

Under the strict constraint of the above operator $\mathcal{O}$, the final canonicalized weights are uniquely determined as $\tilde{\mathbf{A}} = \mathbf{S}^* \mathbf{A}'$ and $\tilde{\mathbf{B}} = \mathbf{B}' \mathbf{S}^*$, thereby collapsing the originally divergent parameter space into a highly deterministic single manifold. Finally, global standardization is performed on the structured matrices to smooth out scale differences across neural network layers. At this point, both the conditional input semantics in Section~\ref{subsec:robust_condition_encoder} and the target parameter manifold possess uniqueness and smoothness, which provides an ideal joint mapping space for the diffusion backbone to perform topology-aware denoising generation in the next subsection.

\begin{figure}[htbp]
    \centering
    \includegraphics[width=1.0\columnwidth]{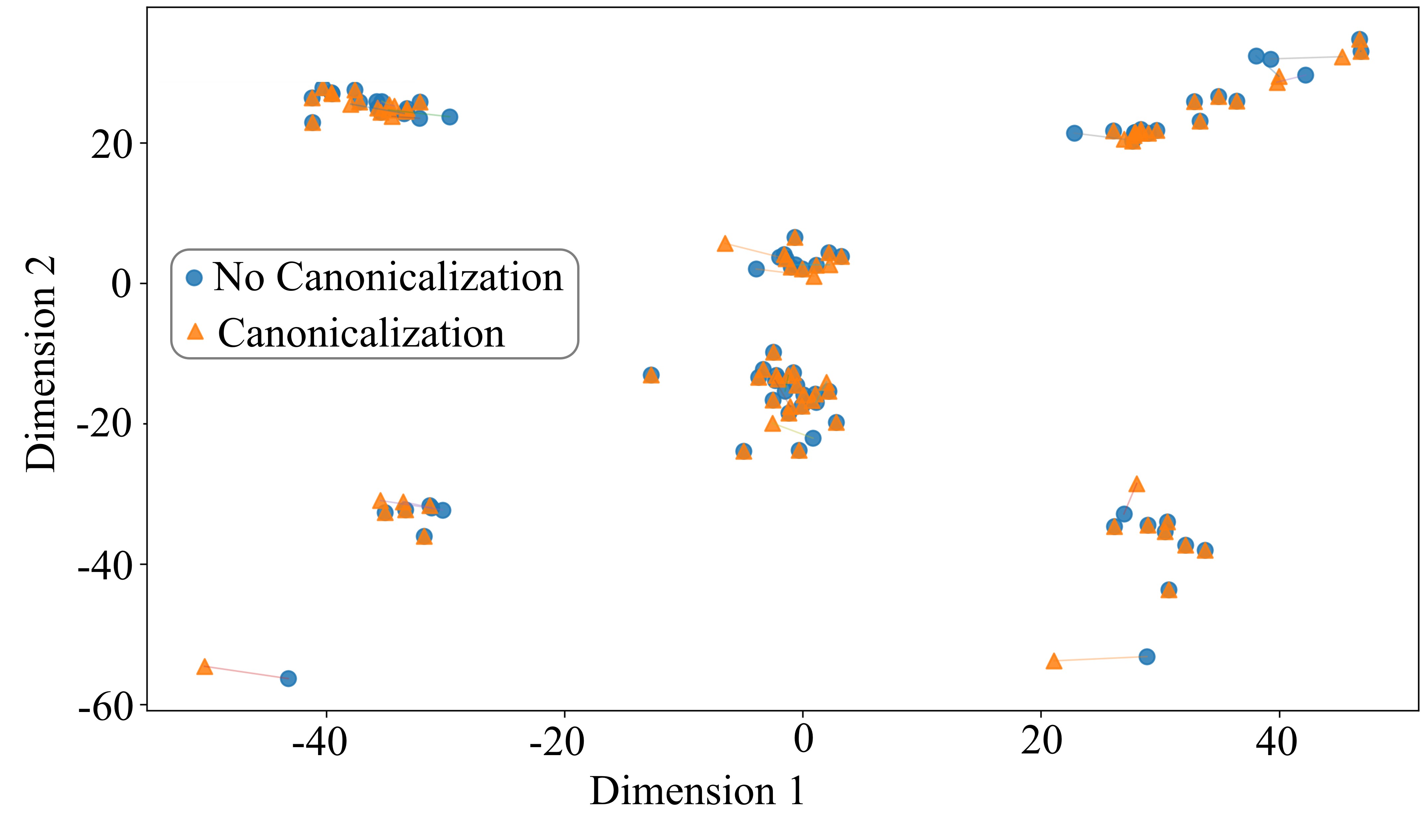}
    \caption{Visualization of the parameter manifold before and after the energy-driven canonicalization operator, projected onto a 2D plane via Principal Component Analysis (PCA). The close overlap between the two point sets indicates that canonicalization resolves equivalent parameter coordinates without disrupting the intrinsic manifold structure.}
    \label{fig:canonicalization_manifold}
\end{figure}

To further verify that the proposed operator does not distort the target parameter manifold, Fig.~\ref{fig:canonicalization_manifold} visualizes the LoRA parameter embeddings before and after canonicalization for one randomly selected seed. By applying PCA to reduce the high-dimensional parameters to two dimensions, we observe that the original and canonicalized points remain tightly aligned across the manifold, with only small local displacements. This observation suggests that the operator mainly removes permutation and sign ambiguities in the coordinate representation while preserving the global geometric structure essential for the generative model's learning process. This paper randomly selects 10 seeds for this manifold analysis; the complete visualizations and quantitative statistics are provided in the supplementary material.

\begin{algorithm}[!t]
\caption{Energy-Driven Canonicalization Operator $\mathcal{O}$}
\label{alg:canonicalization}
\begin{algorithmic}[1]

\REQUIRE LoRA matrices $\mathbf{A}\in\mathbb{R}^{r\times d_{\mathrm{in}}}$ and $\mathbf{B}\in\mathbb{R}^{d_{\mathrm{out}}\times r}$.
\ENSURE Canonicalized matrices $\tilde{\mathbf{A}}$ and $\tilde{\mathbf{B}}$.

\STATE \textit{Stage 1: permutation canonicalization}
\FOR{$i=1,\ldots,r$}
    \STATE $e_i \leftarrow \|\mathbf{A}_{i,:}\|_2 \cdot \|\mathbf{B}_{:,i}\|_2$
\ENDFOR

\STATE $\pi \leftarrow \operatorname{argsort}_{\downarrow}(\{e_i\}_{i=1}^{r})$ with index-based tie breaking
\STATE $\mathbf{P}^{*} \leftarrow \operatorname{Perm}(\pi)$
\STATE $\mathbf{A}' \leftarrow {\mathbf{P}^{*}}^{T}\mathbf{A}$,\quad
$\mathbf{B}' \leftarrow \mathbf{B}\mathbf{P}^{*}$

\STATE \textit{Stage 2: sign canonicalization}
\FOR{$i=1,\ldots,r$}
    \STATE $j_i^* \leftarrow \arg\max_j |\mathbf{A}'_{i,j}|$
    \STATE $s_i \leftarrow \operatorname{sgn}(\mathbf{A}'_{i,j_i^*})$ with $\operatorname{sgn}(0)=1$
\ENDFOR

\STATE $\mathbf{S}^{*} \leftarrow \operatorname{diag}(s_1,\ldots,s_r)$
\STATE $\tilde{\mathbf{A}} \leftarrow \mathbf{S}^{*}\mathbf{A}'$,\quad
$\tilde{\mathbf{B}} \leftarrow \mathbf{B}'\mathbf{S}^{*}$

\STATE \textbf{return} $\tilde{\mathbf{A}},\tilde{\mathbf{B}}$

\end{algorithmic}
\end{algorithm}

\subsection{Structure-Aware Diffusion Backbone}
\label{subsec:method_generation}

After obtaining the canonicalized target manifold $(\tilde{\mathbf{A}}, \tilde{\mathbf{B}})$ and the compact scene semantics $\mathbf{c}$, the final stage of the conditional generation pipeline is the denoising network $\boldsymbol{\epsilon}_\theta$ driven by the Diffusion Transformer (DiT). However, forcibly flattening multi-layer network weights completely destroys their inherent physical topology, and the parameter scales of the LoRA down-projection and up-projection matrices are highly imbalanced. Conventional diffusion backbones combined with global mean squared error (MSE) losses can easily lead to oscillatory optimization trajectories or mode collapse. Therefore, CCPG performs an end-to-end structural reconstruction of the backbone input representation, the conditioning injection paradigm, and the objective function.

\textbf{Topology-Aware Weight Reshaping:} Since parameters from different network layers have different dimensions, this paper partitions the continuous flattened weights into non-overlapping chunks of fixed length, thereby discretizing them into a token sequence $\mathbf{w}_t$ containing $S$ elements. To overcome the structural blind spot of a one-dimensional sequence, in addition to the standard absolute positional encoding $\mathbf{P}_{\mathrm{pos}}$, we further inject a layer-index embedding $\mathbf{P}_{\mathrm{layer}}$ and a matrix-type embedding $\mathbf{P}_{\mathrm{matrix}}$ for each token. The initial hidden representation $\mathbf{T}_0$ at diffusion timestep $t$ is mapped as:
\begin{equation}
    \label{eq:topology_aware_embedding}
    \mathbf{T}_0 = \operatorname{Linear}(\mathbf{w}_t) + \mathbf{P}_{\mathrm{pos}} + \mathbf{P}_{\mathrm{layer}} + \mathbf{P}_{\mathrm{matrix}}
\end{equation}
This joint-awareness mechanism explicitly injects physical coordinates into the hidden space, enabling DiT to accurately capture hierarchical constraints in network weights.

\textbf{Decoupled Spatio-Temporal Conditioning:} In the core network layers of CCPG, the diffusion timestep $t$ and scene semantics $\mathbf{c}$ are injected in a strictly decoupled manner:
\begin{equation}
    \label{eq:decoupled_conditioning}
    \begin{aligned}
        \mathbf{T}' &= \mathbf{T} + \operatorname{AdaLNZero}(\mathbf{T}, t) \odot \operatorname{SelfAttn}(\mathbf{T}) \\
        \mathbf{T}'' &= \mathbf{T}' + \operatorname{CrossAttn}(\mathbf{T}', \mathbf{c}) 
    \end{aligned}
\end{equation}
where $\mathbf{T}$ denotes the input hidden state of the current layer. The timestep $t$ is independently fed into adaptive layer normalization, namely AdaLN-Zero, to globally modulate the denoising trajectory. Its terminal weights are strictly initialized to zero, making the initial state equivalent to an identity function and thereby fundamentally avoiding gradient explosion in the early high-noise regime. Meanwhile, the topology-aware denoising feature $\mathbf{T}'$ serves as the query and actively retrieves environmental semantics $\mathbf{c}$ through cross-attention, enabling precise cross-domain guidance from scenario-specific knowledge to adaptation weights.

\textbf{Size-Aware Compound Optimization:} During optimization, the parameter counts of the down-projection matrix $\tilde{\mathbf{A}}$ and the up-projection matrix $\tilde{\mathbf{B}}$ may differ by several orders of magnitude. From a physical perspective, $\tilde{\mathbf{A}}$ anchors the intrinsic feature subspace of the scenario, and its reconstruction accuracy directly affects the attainable cross-domain adaptation performance. If a global loss is computed without distinction, the reconstruction error of the larger matrix will dominate the gradients, causing the features of the core small matrix to be overwhelmed. To address this issue, this paper leverages the previously introduced topological structure index, namely the Matrix ID, to construct a size-aware weighting mask $\mathbf{\Omega}$, assigning a significantly larger weighting ratio $\rho$ to the down-projection matrix $\tilde{\mathbf{A}}$. Combined with the signal-to-noise-ratio truncation coefficient $\lambda(t) = \min(\operatorname{SNR}(t), \gamma)$~\cite{hang2023efficient} for suppressing optimization oscillations, and the random condition dropout mechanism that supports classifier-free guidance (CFG), the forward diffusion process is defined as:
\begin{equation}
    \label{eq:forward_diffusion}
    \mathbf{w}_t = \sqrt{\bar{\alpha}_t}\mathbf{w}_0 + \sqrt{1-\bar{\alpha}_t}\boldsymbol{\epsilon}, \quad \boldsymbol{\epsilon}\sim\mathcal{N}(\mathbf{0},\mathbf{I}).
\end{equation}
Here, $\mathbf{w}_0$ denotes the clean standardized weight-token sequence and $\bar{\alpha}_t$ denotes the cumulative noise schedule. The final compound optimization objective of CCPG is then formulated as:
\begin{equation}
    \label{eq:dpmg_loss}
    \mathcal{L}_{\mathrm{CCPG}} = \mathbb{E}_{\mathbf{w}_0, \boldsymbol{\epsilon}, \mathbf{c}, t} \left[ \lambda(t) \cdot \left\| \mathbf{\Omega} \odot \left( \boldsymbol{\epsilon} - \boldsymbol{\epsilon}_\theta(\mathbf{w}_t, t, \tilde{\mathbf{c}}) \right) \right\|_2^2 \right]
\end{equation}
where $\mathbf{\Omega}$ has the same token shape as $\mathbf{w}_t$ and assigns weighting ratio $\rho$ to tokens from $\tilde{\mathbf{A}}$, $\lambda(t)=\min(\operatorname{SNR}(t),\gamma)$ is the truncated SNR weight with threshold $\gamma$, and $\tilde{\mathbf{c}}=\mathbf{c}$ with probability $1-p_{\mathrm{drop}}$ and $\tilde{\mathbf{c}}=\emptyset$ with probability $p_{\mathrm{drop}}$. This objective function forces the model to prioritize the alignment of the core physical basis of the target environment while maintaining stable convergence.

Through parameter chunking, decoupled condition injection, and size-aware optimization, CCPG successfully transforms the complex scalar weight fitting problem into a manifold denoising task equipped with rigorous physical intuition and topological awareness. During inference, this complete generation pipeline can achieve high-fidelity generation of core weights for unseen scenarios with extremely low computational overhead at terminal devices.

\section{Experiments}

\subsection{Datasets and Metrics}

The proposed CCPG is evaluated on two standard wireless communication datasets, DeepMIMO~\cite{alkhateeb2019deepmimo} and WAIR-D~\cite{wang2022waird}. The experiments cover two representative physical-layer tasks: channel estimation (CE) and CSI compression feedback (CF). Specifically, experiments are conducted on DeepMIMO CF, DeepMIMO CE, and WAIR-D CF to examine the training-free adaptation ability of CCPG under different channel distributions and task formulations.

For the CF task, the compression ratio is set to $\mathrm{CR}=1/4$ throughout the experiments. This setting is commonly adopted in CSI feedback studies and provides a representative trade-off between feedback overhead and reconstruction difficulty. For the CE task, noisy channel observations are generated under $\mathrm{SNR}=0~\mathrm{dB}$, which corresponds to a challenging low-SNR setting and is suitable for evaluating model robustness under degraded channel observations.

Unless otherwise specified, these task-specific configurations are used consistently for all methods to ensure fair comparisons. The main paper focuses on these representative settings, while additional results under other compression ratios and SNR levels will be included in supplementary materials.

Following common practice in CSI reconstruction and channel estimation, the normalized mean squared error (NMSE) in dB is adopted as the primary evaluation metric:
\begin{equation}
    \label{eq:nmse}
    \mathrm{NMSE}
    =
    10\log_{10}
    \left(
    \frac{\|\mathbf{H}_{\mathrm{out}}-\mathbf{H}\|_F^2}
    {\|\mathbf{H}\|_F^2}
    \right),
\end{equation}
where $\mathbf{H}$ denotes the ground-truth channel and $\mathbf{H}_{\mathrm{out}}$ denotes the reconstructed or estimated channel. A lower NMSE indicates better reconstruction accuracy.

\subsection{Implementation Details}
\label{subsec:implementation_details}

\textbf{Meta-dataset construction.}
To construct the condition-parameter meta-dataset, $10\%$ of the scenarios are first randomly sampled to pre-train a backbone Transformer-based model as the base model. Once pre-trained, the backbone model is frozen. LoRA modules inserted into the decoder FFN layers are then optimized across the remaining $90\%$ of independent scenarios. The resulting $\langle$condition matrix, adaptation weight$\rangle$ pairs are collected as meta-samples. These samples are randomly split into training, validation, and test sets with a ratio of $8:1:1$, ensuring that the final evaluation directly measures the cross-scenario parameter generation capability of CCPG without any online gradient updates.

\textbf{Architecture configurations.}
All conditional parameter generation pipelines are implemented and evaluated on a single NVIDIA H100 GPU. In the condition extraction stage, the truncation hyperparameters of the physics-aware dimensionality reduction matrix $\mathbf{C}_{\mathrm{svd}}$ are set to $L=224$ and $d=512$. The subsequent Perceiver Resampler has a hidden dimension of $1024$ and outputs a fixed-length latent condition sequence with $M=128$ tokens. In the parameter generation stage, the topology-aware denoising backbone is instantiated as a DiT model with $12$ Transformer blocks and $4$ attention heads.

\textbf{Training and inference setup.}
During end-to-end training, CCPG is trained for $5000$ epochs with a batch size of $64$. AdamW is used as the optimizer with a weight decay of $0.05$. The maximum learning rate is set to $3\times 10^{-4}$ and is scheduled using a warmup strategy followed by cosine annealing. To support classifier-free guidance (CFG) and improve conditional generation robustness, the condition dropout probability $p_{\mathrm{drop}}$ is set to $0.1$. In addition, an exponential moving average (EMA) with a decay rate of $0.999$ is applied to stabilize the denoising network. During training-free online deployment, CCPG uses DDIM sampling~\cite{song2021ddim} with $50$ denoising steps to generate the target LoRA parameters.

\subsection{Baselines}
To evaluate the effectiveness of CCPG comprehensively, it is compared with both adaptation references and existing parameter generation baselines.

First, the frozen base encoder-decoder is used as the \textbf{Base}. This model is pre-trained only on the $10\%$ source-domain scenarios and is directly evaluated on unseen test scenarios without any parameter update. Therefore, it reflects the performance degradation caused by domain shifts when a static model is deployed in heterogeneous wireless scenarios.

Second, \textbf{Online Training} is introduced as an empirical online adaptation reference. For each target test scenario, the LoRA modules are optimized for $200$ epochs using samples from the corresponding target scenario. This setting represents a strong but computationally expensive adaptation scheme. It should be noted that this reference is not a strict mathematical upper bound, since its performance is still affected by the finite number of optimization steps, available target samples, and training hyperparameters. Nevertheless, it provides a useful empirical reference for the accuracy that conventional online gradient-based adaptation can achieve.

Finally, CCPG is compared with several representative generative parameter generation methods, including W-Diff~\cite{xie2024wdiff}, LoRAGen~\cite{huang2026loragen}, D2NWG~\cite{ICLR2025_f74d7957}, and MetaDiff~\cite{zhang2024metadiff}. For fair comparison, all generative baselines are implemented based on their official code repositories, with only necessary modifications for the experimental setting of this paper. Specifically, they are adapted to the same meta-dataset, use the same training/validation/test split, and generate the same target LoRA modules as CCPG.

\subsection{Main Results}
\label{subsec:main_results}

The main performance results are summarized in Table~\ref{tab:performance_comparison}. For compact presentation, the table reports three representative scenarios for each dataset-task pair, while the ``Average'' column is computed over all 10 held-out test scenarios. The complete scenario-wise performance and time-consumption results are provided in the supplementary material.

As shown in Table~\ref{tab:performance_comparison}, the frozen base model suffers from severe degradation on unseen scenarios, indicating that a static model is insufficient under strong environmental heterogeneity. Existing generative baselines also fail to provide reliable adaptation in this communication-oriented setting. In several cases, their generated parameters even cause negative transfer and perform worse than the frozen base model. This suggests that parameter generation methods developed for generic neural network weight distributions cannot be directly transferred to physical-layer scenarios, where condition inputs are high-dimensional, noisy, and strongly constrained by channel structures.

In contrast, CCPG consistently achieves the best performance among all training-free generative methods. On average, it improves over the strongest generative baseline by $3.41$ dB on DeepMIMO CF, $2.44$ dB on DeepMIMO CE, and $4.77$ dB on WAIR-D CF. These gains demonstrate the necessity of physics-aware condition reduction, topology-preserving parameter representation, and diffusion-based structured denoising for reliable LoRA weight generation in heterogeneous wireless scenarios.

The complete scenario-wise results further confirm the stability of CCPG. On DeepMIMO CF, CCPG remains close to Online Training across all 10 scenarios, rather than relying on a few favorable cases. A similar trend is observed on DeepMIMO CE, where noisy observations make the task more sensitive and the absolute NMSE margin slightly larger. This indicates that CCPG learns an effective condition-dependent mapping across different physical-layer objectives. The WAIR-D CF results are particularly informative: under more irregular real-world scenario distributions, several baselines become unstable and occasionally fall below the frozen Base model, whereas CCPG consistently recovers most of the gap between Base and Online Training. This suggests that preserving parameter topology and aligning LoRA weight ambiguities become even more critical when channel distributions are less regular and more environment-dependent.

More importantly, CCPG approaches the empirical online training reference without requiring any online gradient update. For example, on DeepMIMO CF, it achieves an average NMSE of $-26.96$ dB, close to the Online Training result of $-27.22$ dB. In the 94\_3p5 scenario, CCPG even slightly outperforms the finite-step reference, which does not imply superiority over the true optimum, but shows that the generated parameters can serve as a strong ready-to-use adaptation solution when online training is constrained by optimization dynamics or limited target-scenario samples.

Therefore, CCPG achieves near online training performance while reducing online adaptation time from minutes to seconds, making it more suitable for practical training-free deployment in dynamic wireless communication systems.

The time comparison in Table~\ref{tab:time_comparison_formatted} further highlights the deployment advantage of CCPG. Online training requires 200 epochs of online optimization for each target scenario, resulting in an average adaptation time of $45$--$61$ minutes depending on the task. By contrast, CCPG only requires a single conditional denoising process and completes scenario adaptation within approximately $3$ seconds.

\begin{table*}[htbp]
  \caption{Performance comparison of different methods across various scenarios. For each dataset-task pair, three scenario IDs are randomly sampled from 10 held-out test scenarios for compact presentation. The ``Average'' column is computed over all 10 held-out test scenarios, and the complete results are provided in the supplementary material. Lower NMSE (dB) indicates better performance.}
  \label{tab:performance_comparison}
  \centering
  \resizebox{\textwidth}{!}{%
    \begin{tabular}{l cccc cccc cccc}
      \toprule
      & \multicolumn{4}{c}{DeepMIMO CF} 
      & \multicolumn{4}{c}{DeepMIMO CE} 
      & \multicolumn{4}{c}{WAIR-D CF} \\
      \cmidrule(lr){2-5} 
      \cmidrule(lr){6-9} 
      \cmidrule(lr){10-13}
      Method / Scen\_id 
      & 4\_3p5 & 14\_28 & 94\_3p5 & Average 
      & 4\_3p5 & 14\_28 & 94\_3p5 & Average 
      & 09493 & 09513 & 09957 & Average \\
      \midrule
      Base
      & -20.36 & -21.01 & -21.61 & -21.24 
      & -17.92 & -16.77 & -17.35 & -17.04 
      & -17.94 & -17.01 & -18.19 & -17.08 \\
      
      Online Training
      & \textbf{-27.56} & \textbf{-27.89} & \textbf{-28.07} & \textbf{-27.22} 
      & \textbf{-22.36} & \textbf{-21.89} & \textbf{-22.28} & \textbf{-21.97} 
      & \textbf{-20.79} & \textbf{-22.04} & \textbf{-22.00} & \textbf{-20.87} \\
      
      W-Diff 
      & -21.63 & -20.57 & -21.19 & -22.33 
      & -18.63 & -17.77 & -15.29 & -16.04 
      & -15.78 & -16.43 & -12.26 & -14.71 \\
      
      D2NWG 
      & -19.36 & -21.09 & -18.54 & -19.88 
      & -13.46 & -16.32 & -16.91 & -15.42 
      & -14.81 & -11.90 & -13.64 & -14.40 \\
      
      LoRAGen 
      & -22.09 & -18.23 & -19.33 & -19.21 
      & -17.62 & -17.04 & -18.94 & -17.70 
      & -16.98 & -12.38 & -15.06 & -15.22 \\
      
      MetaDiff 
      & -23.00 & -24.28 & -23.94 & -23.55 
      & -18.92 & -18.43 & -19.19 & -18.96 
      & -16.04 & -13.88 & -16.13 & -15.82 \\
      \midrule
      \textbf{CCPG (Ours)} 
      & \textbf{-27.44} & \textbf{-27.88} & \textbf{-28.10} & \textbf{-26.96} 
      & \textbf{-22.01} & \textbf{-21.33} & \textbf{-22.16} & \textbf{-21.40} 
      & \textbf{-20.57} & \textbf{-22.10} & \textbf{-21.86} & \textbf{-20.59} \\
      \bottomrule
    \end{tabular}%
  }
\end{table*}

\begin{table*}[htbp]
\caption{Adaptation-time comparison across various scenarios. The three listed scenario IDs are randomly sampled from the test scenarios for compact presentation. The ``Average'' column is computed over all held-out test scenarios, and the complete scenario-wise results are provided in the supplementary material. Bold numbers indicate the results of CCPG.}
  \label{tab:time_comparison_formatted}
  \centering
  \resizebox{\textwidth}{!}{%
    \begin{tabular}{l cccc cccc cccc}
      \toprule
      & \multicolumn{4}{c}{DeepMIMO CF} 
      & \multicolumn{4}{c}{DeepMIMO CE} 
      & \multicolumn{4}{c}{WAIR-D CF} \\
      \cmidrule(lr){2-5} 
      \cmidrule(lr){6-9} 
      \cmidrule(lr){10-13}
      Method / Scen\_id 
      & 4\_3p5 & 14\_28 & 94\_3p5 & Average 
      & 4\_3p5 & 14\_28 & 94\_3p5 & Average 
      & 09493 & 09513 & 09957 & Average \\
      \midrule
      Online Training
      & 61 min & 62 min & 60 min & 60 min 
      & 53 min & 43 min & 36 min & 45 min 
      & 62 min & 63 min & 62 min & 61 min \\
      
      \textbf{CCPG (Ours)} 
      & \textbf{3.0 s} & \textbf{3.5 s} & \textbf{4.7 s} & \textbf{3.2 s} 
      & \textbf{2.2 s} & \textbf{1.8 s} & \textbf{3.5 s} & \textbf{2.8 s} 
      & \textbf{3.2 s} & \textbf{4.3 s} & \textbf{3.7 s} & \textbf{3.5 s} \\
      \bottomrule
    \end{tabular}%
  }
\end{table*}

\subsection{Ablation Studies}
\label{subsec:ablation}

To further investigate the independent contribution of each core component of CCPG to cross-domain generalization, controlled leave-one-out ablation studies are conducted. Table~\ref{tab:performance_ablation} reports the performance degradation of different model variants on the test set after four key mechanisms are removed, with the complete scenario-wise ablation tables provided in the supplementary material.

\textbf{Effect of parameter canonicalization (no-canonicalization):} 
The role of parameter canonicalization and alignment is first examined. When training is performed using raw, non-canonicalized data, the generation network suffers from extremely severe mode collapse. As shown in the table, the average NMSE of this variant on the DeepMIMO CF task drops sharply from $-26.96$ dB to $-8.10$ dB, resulting in a performance loss of nearly $20$ dB. The underlying reason is that, without imposing uniqueness constraints on the target parameter manifold, the highly ill-posed solution space forces the diffusion model to average multiple parameter solutions that are geometrically distinct in the coordinate space but equivalent in function. This inevitably leads to the generation of smoothed noise without physical meaning.

\textbf{Effect of semantic resampling (no-resampler):} 
To verify the effectiveness of condition preprocessing, the Perceiver Resampler is replaced with global average pooling followed by a multilayer perceptron. The results show that this replacement causes severe performance degradation, with the average NMSE dropping by 11 to 14 dB across different tasks. Simply applying global pooling to high-dimensional CSI signals indiscriminately mixes different physical eigenmodes, resulting in excessive semantic smoothing. In contrast, CCPG employs the resampler as an active query mechanism, effectively decoupling channel complexity from generation difficulty.

\textbf{Effect of structural topology embedding (no-structure-aware):} 
When the Layer ID and Matrix ID embeddings are removed, the model performance decreases substantially, with average performance losses of 7.4 to 8.7 dB across tasks. This phenomenon confirms that a plain DiT architecture lacks topological intuition when processing neural network weights. Without these physical position identifiers, the network cannot distinguish the semantic hierarchy at different network depths or the heterogeneity between $\mathbf{A}$ and $\mathbf{B}$ matrices, causing cross-layer parameter generation to become disordered. This verifies the necessity of explicitly injecting weight topology priors.

\textbf{Mechanism of asymmetric size awareness (no-size-aware):} 
Finally, the size-aware weighting mask is removed from the optimization objective, and the mean squared error is directly computed over the entire weight sequence. In addition, Fig.~\ref{fig:line_chart_r} further illustrates the influence of different asymmetric weighting ratios $\rho$ on reconstruction accuracy. Without any explicit weighting, the implicit gradient update ratio is equivalent to the parameter-count ratio between the two matrices, in which case the natural weight of $\mathbf{A}$ is only approximately 0.03. The experimental results show that the network achieves the best generalization performance when $\rho=0.3$, whereas using equal weights ($\rho=0.5$) causes a performance drop of approximately 1.8 to 3.3 dB. This empirical observation reveals the intrinsic asymmetric physical mechanism within the LoRA architecture: although the down-projection matrix $\mathbf{A}$ contains fewer parameters, it anchors the intrinsic feature subspace of the scenario. Without an asymmetric mask, the large gradients produced by the up-projection matrix $\mathbf{B}$ overwhelm the optimization trajectory of $\mathbf{A}$, leading to a shift in the physical basis of channel reconstruction.

The above analysis demonstrates that the superior performance of CCPG is built upon the deep coupling of physical channel feature compression, weight topology awareness, and asymmetric manifold optimization.

\begin{table*}[htbp]
  \caption{Performance comparison of different ablation variants across various scenarios. For each dataset-task pair, three scenario IDs are randomly sampled from 10 held-out test scenarios for compact presentation. The ``Average'' column is computed over all 10 held-out test scenarios, and the complete scenario-wise ablation results are provided in the supplementary material.}
  \label{tab:performance_ablation}
  \centering
  \resizebox{\textwidth}{!}{%
    \begin{tabular}{l cccc cccc cccc}
      \toprule
      & \multicolumn{4}{c}{DeepMIMO CF} & \multicolumn{4}{c}{DeepMIMO CE} & \multicolumn{4}{c}{WAIR-D CF} \\
      \cmidrule(lr){2-5} \cmidrule(lr){6-9} \cmidrule(lr){10-13}
      Method / Scen\_id             & 4\_3p5 & 14\_28 & 94\_3p5 & Average & 4\_3p5 & 14\_28 & 94\_3p5 & Average & 09493 & 09513 & 09957 & Average \\
      \midrule
      no-canonicalization   & -8.27        & -6.41        & -6.72        & -8.10        & -3.22        & -4.66        & -4.08        & -4.47        & -1.04        & -2.55        & 0.96        & -0.15        \\
      no-resampler          & -13.69       & -14.22       & -10.37       & -14.58       & -11.58       & -9.30        & -9.08        & -9.93        & -9.34        & -10.00       & -8.51       & -7.05        \\
      no-structure-aware    & -18.65       & -18.20       & -19.09       & -18.89       & -10.32       & -15.08       & -14.33       & -13.31       & -13.64       & -13.86       & -11.12       & -12.22       \\
      no-size-aware         & -25.22       & -24.87       & -25.00       & -24.99       & -19.39       & -18.76       & -19.41       & -18.79       & -17.21       & -18.56       & -16.99       & -17.32       \\
      \midrule
      \textbf{CCPG (Ours)}& \textbf{-27.46}& \textbf{-27.85}& \textbf{-28.10}& \textbf{-26.96}& \textbf{-22.01}& \textbf{-21.33}& \textbf{-22.16}& \textbf{-21.40}& \textbf{-20.57}& \textbf{-22.10}& \textbf{-21.86}& \textbf{-21.13}\\
      \bottomrule
    \end{tabular}%
  }
\end{table*}

\pgfplotsset{compat=1.18}

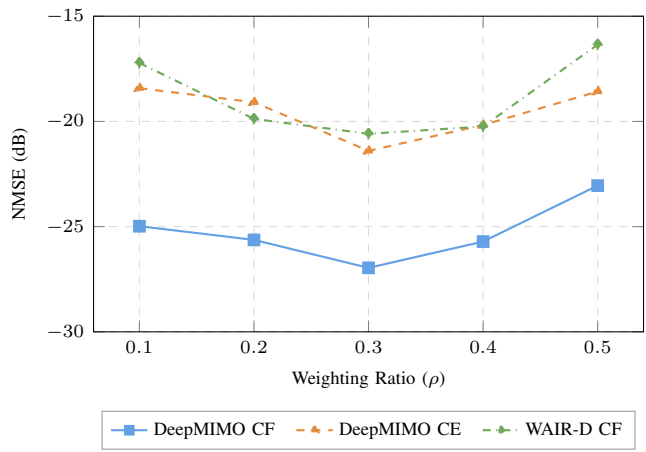
\begin{figure}[htbp]
    \centering

    \pgfplotsset{compat=1.18}

    \definecolor{color1}{RGB}{100,160,230} 
    \definecolor{color2}{RGB}{230,140,60} 
    \definecolor{color3}{RGB}{110,170,90} 

    \begin{tikzpicture}
        \begin{axis}[
            width=1.0\columnwidth,
            height=0.65\columnwidth,
            xlabel={Weighting Ratio ($\rho$)},
            ylabel={NMSE (dB)},
            xtick={0.1, 0.2, 0.3, 0.4, 0.5},
            grid=major,
            grid style={dashed, gray!30},
            ymin=-30, ymax=-15,
            label style={font=\scriptsize},
            tick label style={font=\scriptsize},
            axis line style={black},
            legend style={
                at={(0.5,-0.25)},
                anchor=north,
                legend columns=3, 
                font=\scriptsize,
                draw=gray!80,
                fill=white,
                /tikz/every even column/.append style={column sep=0.2cm}
            },
            legend image post style={scale=0.8},
        ]

        \addplot[
            color=color1, 
            mark=square*,
            thick,
        ] coordinates {
            (0.1, -24.98) (0.2, -25.63) (0.3, -26.96) (0.4, -25.71) (0.5, -23.05)
        };
        \addlegendentry{DeepMIMO CF}

        \addplot[
            color=color2,
            mark=triangle*,
            dashed,
            thick,
        ] coordinates {
            (0.1, -18.42) (0.2, -19.09) (0.3, -21.40) (0.4, -20.17) (0.5, -18.58)
        };
        \addlegendentry{DeepMIMO CE}

        \addplot[
            color=color3,
            mark=diamond*,
            dashdotted,
            thick,
        ] coordinates {
            (0.1, -17.22) (0.2, -19.88) (0.3, -20.59) (0.4, -20.24) (0.5, -16.35)
        };
        \addlegendentry{WAIR-D CF}

        \end{axis}
    \end{tikzpicture}
    \caption{Performance curve of average NMSE with respect to the asymmetric weighting ratio $\rho$.}
    \label{fig:line_chart_r}
\end{figure}

\section{Limitations}
\label{sec:limitation}

Although the proposed CCPG achieves promising results in training-free cross-scenario adaptation, several aspects remain to be improved. On the one hand, the current experiments are mainly conducted on the DeepMIMO and WAIR-D datasets. Although these datasets cover representative wireless propagation scenarios, evaluation under more complex real-world conditions, such as high mobility, severe blockage, and multi-cell interference, remains insufficient. On the other hand, this paper mainly focuses on the lightweight generation of the core bottleneck parameters in the decoder and adopts a fixed LoRA rank and structured partitioning strategy, whereas adaptive parameter configuration under different network scales has not yet been fully explored.

Future work will introduce more real-world measured channel data and dynamic communication scenarios to conduct a more comprehensive evaluation of the generalization stability of CCPG. Adaptive bottleneck parameter selection, LoRA rank configuration, and weight structure partitioning strategies will also be further investigated to improve the applicability and deployment flexibility of the proposed method in practical wireless communication systems.

\section{Conclusion}
\label{sec:conclusion}

This paper addresses the cross-scenario adaptation problem of massive MIMO systems in complex heterogeneous scenarios and proposes CCPG, a training-free conditional parameter generation architecture. To overcome the computational burden of conventional online training at terminal devices and the topological collapse problem in standard parameter generation methods, this paper localizes the generalization target to lightweight LoRA matrices in the decoder FFN layers and integrates three core mechanisms: channel condition compression, energy-driven parameter-space canonicalization, and a structure-aware diffusion backbone. Evaluations on the DeepMIMO and WAIR-D datasets demonstrate that CCPG achieves strong adaptation capability for unseen scenarios. In terms of communication reconstruction accuracy, it closely approaches the empirical online training reference while reducing scenario adaptation time at the terminal side to approximately 3 seconds. This study demonstrates the potential of replacing conventional online adaptation with structure-aware parameter generation, providing a feasible and general solution for large-scale, low-cost, plug-and-play dynamic physical-layer deployment in next-generation intelligent 6G communication networks.

\bibliographystyle{IEEEtran}
\bibliography{references}

\end{document}